\documentclass[sigconf]{acmart-me}

\usepackage{booktabs} 
\usepackage{url}
\usepackage{color}
\usepackage{enumitem}
\usepackage{subcaption}
\usepackage{tabularx}
\usepackage{multirow}
\usepackage{hyperref}
\usepackage{microtype}
\setcopyright{rightsretained}

\acmDOI{}

\acmISBN{}

\acmConference[MediaEval'19]{Multimedia Evaluation Workshop}{27-29 October 2019}{Sophia Antipolis, France} 
\acmYear{}
\copyrightyear{}

\acmPrice{}

\begin{document}
\title{MediaEval 2019: Concealed FGSM Perturbations for Privacy Preservation}

\author{Panagiotis Linardos, Suzanne Little, Kevin McGuinness}
\affiliation{Dublin City University}
\email{linardos.akis@gmail.com}

%
%
%
%
%

\renewcommand{\shortauthors}{P. Linardos et al.}
\renewcommand{\shorttitle}{Pixel Privacy}

\begin{abstract}
This work tackles the Pixel Privacy task put forth by MediaEval 2019. Our goal is to manipulate images in a way that conceals them from automatic scene classifiers while preserving the original image quality. We use the fast gradient sign method, which normally has a corrupting influence on image appeal, and devise two methods to minimize the damage. The first approach uses a map of pixel locations that are either salient or flat, and directs perturbations away from them. The second approach subtracts the gradient of an aesthetics evaluation model from the gradient of the attack model to guide the perturbations towards a direction that preserves appeal. We make our code available at: \url{https://git.io/JesXr}.
\end{abstract}

%
%
%
%
%


\maketitle

\section{Introduction}
\label{sec:intro}

The Pixel Privacy task, introduced by MediaEval~\cite{mediaeval2019}, aims at developing methods for manipulating images in a way that fools automatic scene classifiers (referred to as attack models). As an added constraint, the images should not exhibit a decrease in aesthetic quality. The organizers made available the Places365-Standard data set~\cite{zhou2017places} along with a pre-trained ResNet~\cite{resnet} attack model for the task.

The contribution of image enhancement techniques in privacy protection has been previously explored~\cite{choi2017geo}, showing that even popular filters used in social media have a cloaking effect against geo-location algorithms. A more recent work by Liu et al.~\cite{liu2018first} proposed a perturbation-based approach (white-box) and a transfer style approach (black-box). Similar to the first module in that work, we propose two perturbation-based approaches and explore ways to localize the perturbations in a manner that does not reduce appeal.
 
\section{Approach}
\label{sec:approach}
We developed two approaches, both of which utilize FGSM ~\cite{goodfellow2014explaining}. FGSM uses the gradient of the attack model and changes the pixel values by nudging them towards the direction that maximizes the loss. Furthermore, the strength of these perturbations varies and is represented by the $\epsilon$ value.  


\begin{figure}
    \centering
    \begin{subfigure}[b]{0.475\linewidth}  
        \centering 
        \includegraphics[width=\linewidth]{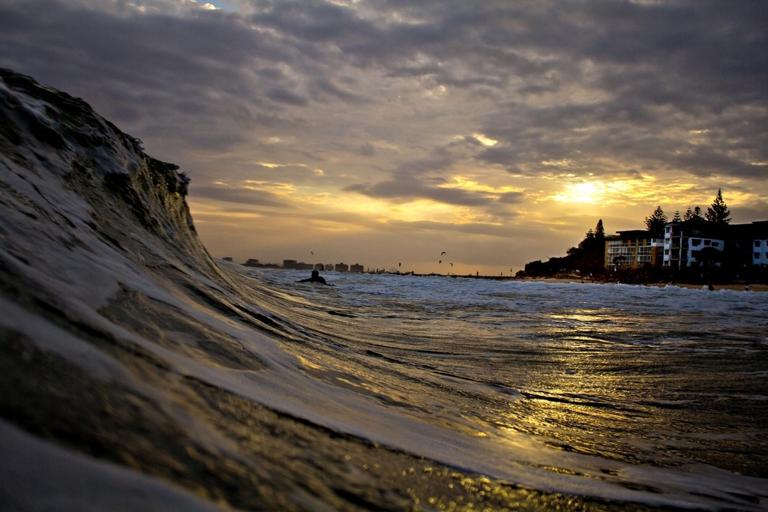}
        \caption[]%
        {{ Original Image}}    
        \label{maps14}
    \end{subfigure}
    \hfill
    \begin{subfigure}[b]{0.475\linewidth}  
        \centering 
        \includegraphics[width=\linewidth]{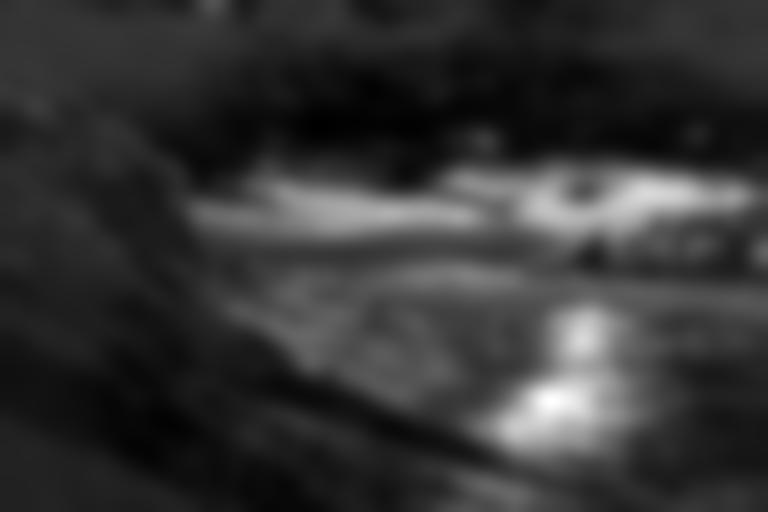}
        \caption[]%
        {{ Sobel Map}}    
        \label{maps24}
    \end{subfigure}
    \vskip\baselineskip
    \begin{subfigure}[b]{0.475\linewidth}   
        \centering 
        \includegraphics[width=\linewidth]{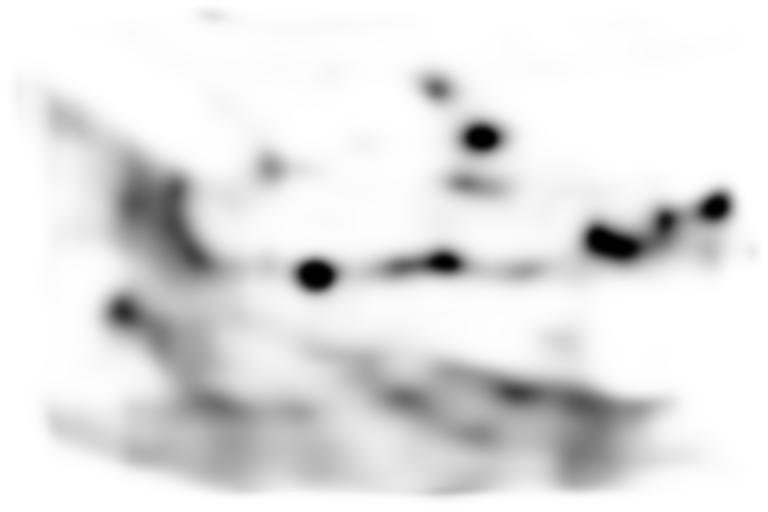}
        \caption[]%
        {{ Reverse Saliency Map}}    
        \label{maps34}
    \end{subfigure}
    \quad
    \begin{subfigure}[b]{0.475\linewidth}   
        \centering 
        \includegraphics[width=\linewidth]{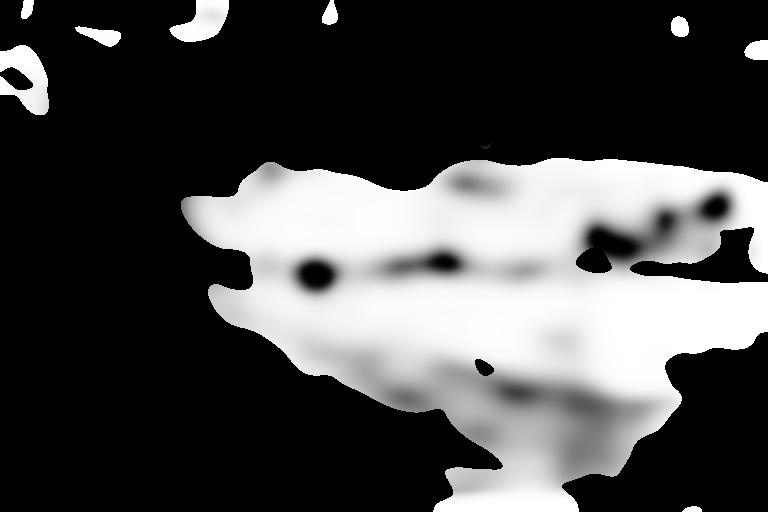}
        \caption[]%
        {{ Final Map}}    
        \label{maps44}
    \end{subfigure}
    \caption[ Maps used to localize perturbations ]
    { Maps used to constrain perturbations on less obvious areas. The reverse saliency map along with the Sobel map produce the final map.} 
    \label{fig:maps}
\end{figure}

\subsection{Salient Defence}
Our first approach combines two maps: one is a measure of saliency and the other a measure of flatness. Salient areas are the ones that are more likely to attract the eye of an observer, and are predicted by a DNN. In particular, we use SalBCE~\cite{linardos2019simple} trained on the SALICON dataset~\cite{jiang2015salicon}. Furthermore, perturbations become more obvious when they are located in flat areas. For this reason, we also used a Sobel filter~\cite{Sobel19683x3}, which detects areas where edges are more prevalent. Gaussian blurring ($\sigma$=10) is applied to spread the detected edges, forming the final Sobel map. 
The saliency map is reversed so that the pixels corresponding to salient areas are zeroed out. Then, pixels where the mean value is below average on the Sobel map (hence more likely to be on a flat area) are also zeroed out. The resulting map \textit{M}, the sign of the network's gradient \textit{g}, and the value $\epsilon$ are multiplied and added to the original image \textit{I}, completing the modification. Figure~\ref{fig:maps} illustrates an example of map generation.
\begin{equation}
I\textsubscript{modified}=M\circ sgn(g(I))\circ\epsilon+I
\end{equation}

Additionally, we used a popular filter for image manipulation, namely tilt-shift to inspect how it affects the efficacy of our approach. Tilt-shift essentially blurs parts of the background while intensifying foreground. In our case we used the saliency maps as an estimate of the foreground to be intensified, blurring the rest.

\subsection{Coupled Optimization}
The second approach exploits the gradients of both the attack model and the aesthetics evaluation algorithm. The aesthetics evaluation in our case is the NIMA algorithm~\cite{talebi2018nima}. Since the networks differ significantly, the gradients are first scaled to be brought to the same range [0,1]. Afterwards, NIMA's gradient is subtracted from ResNet's and as a result we get the sign of the total gradient and multiply that by $\epsilon$:
\begin{equation}
I\textsubscript{modified}= sgn(g\textsubscript{\textit{ResNet}}(I)-g\textsubscript{\textit{NIMA}}(I))\circ\epsilon+I
\end{equation}

\begin{figure}
\includegraphics[width=0.47\linewidth]{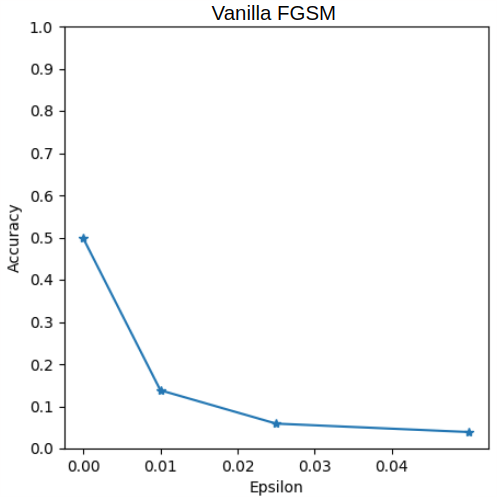} 
\includegraphics[width=0.47\linewidth]{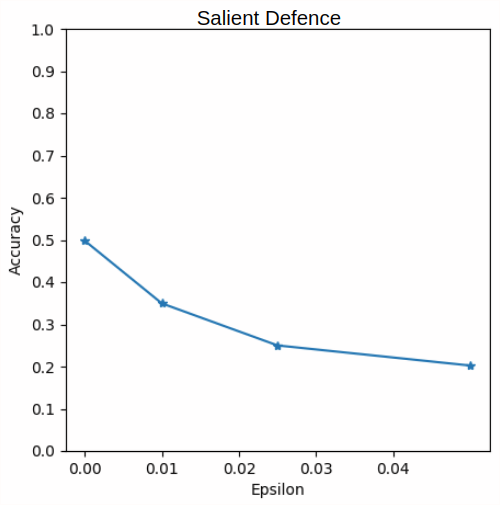} 
\includegraphics[width=0.47\linewidth]{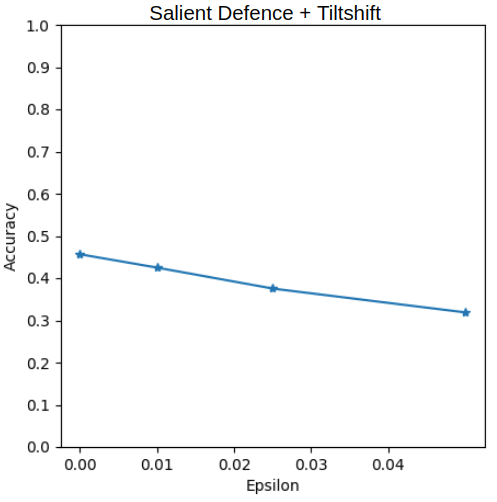}
\includegraphics[width=0.47\linewidth]{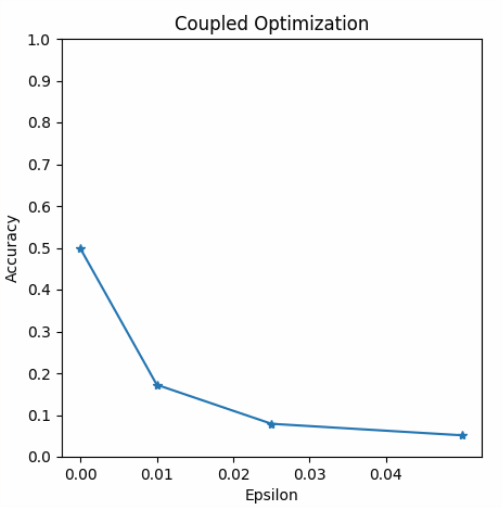}
\caption[ Epsilons ]
    {Attack model accuracy under differing values of $\epsilon$.} 
\label{figepsilons}
\end{figure}

\section{Results and Analysis}
In our initial experiments, we used the full-resolution images from Places365 and applied a variety of $\epsilon$ values to investigate how they affect the accuracy of the attack model (Figure~\ref{figepsilons}). Salient Defence perturbs less pixels, which explains the lower impact on accuracy compared to the vanilla FGSM. We also note that the tilt-shift filter further reduces the efficacy of those perturbations. The coupled optimization approach, has a higher impact on the accuracy of the attack model, as it manipulates all the pixels of the image.

The test set, as evaluated by the MediaEval team (Table~\ref{tab:Results}) was first downsampled to $256\times 256$ and the algorithms were applied afterwards. Note that this set includes only images that ResNet predicts successfully, and so the initial accuracy ($\epsilon = 0$) is 100\%. In that case it seems that the tilt-shift effect actually adds to the efficacy of the perturbations, bringing the accuracy of the attack model down while increasing the aesthetics score. 

To test NIMA's sensitivity to perturbations, we used FGSM (vanilla) with a very high $\epsilon = 0.15$ on a small subset (100) of the validation images. This type of attack effectively ruins the visual appeal; however, the NIMA score drops by only a small amount (from 4.26 to 3.98). This indicates that NIMA has a low-sensitivity to adversarial perturbations. This could be explained by the fact that NIMA was trained on AVA~\cite{AVA}, a dataset collected by photographers. The model is, therefore, sensitive to high-level concepts of aesthetic appeal, such as the rule of thirds, but has not been trained to be sensitive to the low-level corrupting influence of perturbations. 

\begin{figure}
    \centering
    \begin{subfigure}[b]{0.475\linewidth}  
        \centering 
        \includegraphics[width=\linewidth]{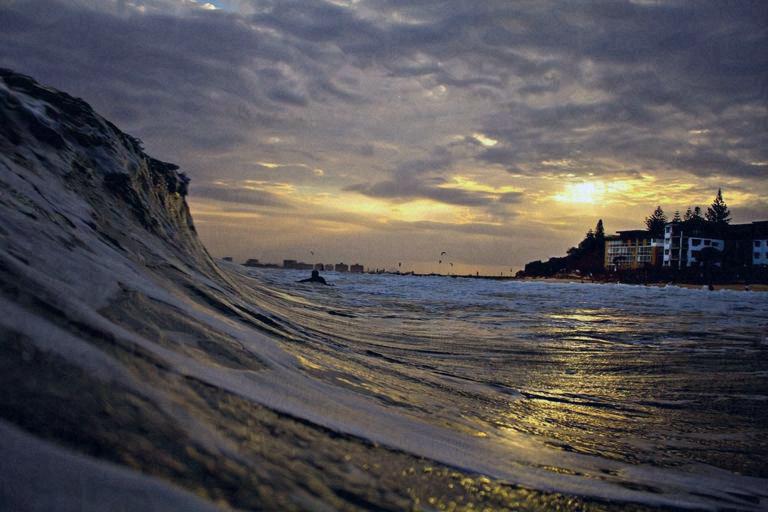}
        \caption[]%
        {{Vanilla FGSM, $\epsilon=0.05$}}    
        \label{maps14}
    \end{subfigure}
    \hfill
    \begin{subfigure}[b]{0.475\linewidth}  
        \centering 
        \includegraphics[width=\linewidth]{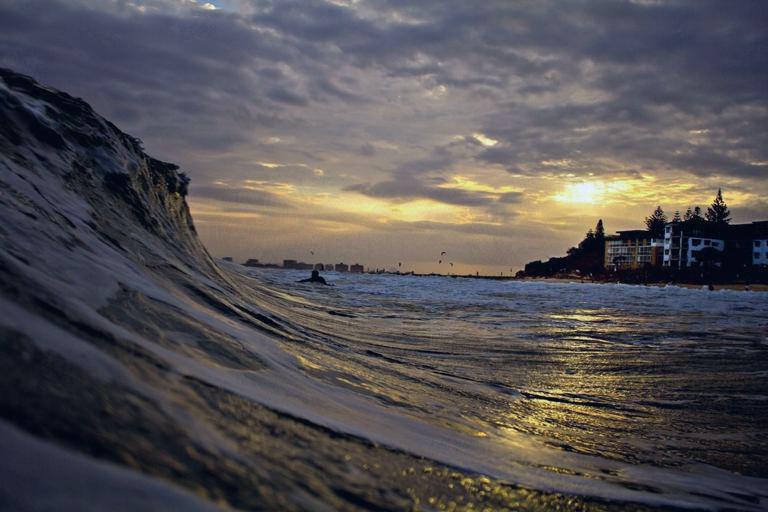}
        \caption[]%
        {{ Salient Defence, $\epsilon=0.05$}}    
        \label{maps24}
    \end{subfigure}
    \vskip\baselineskip
    \begin{subfigure}[b]{0.475\linewidth}   
        \centering 
        \includegraphics[width=\linewidth]{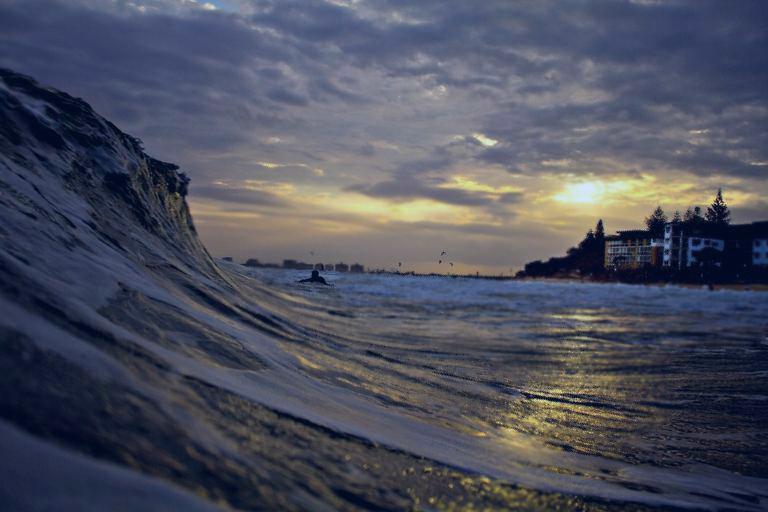}
        \caption[]%
        {{Salient Defence \& tshift, $\epsilon=0.01$}}    
        \label{maps34}
    \end{subfigure}
    \quad
    \begin{subfigure}[b]{0.475\linewidth}   
        \centering 
        \includegraphics[width=\linewidth]{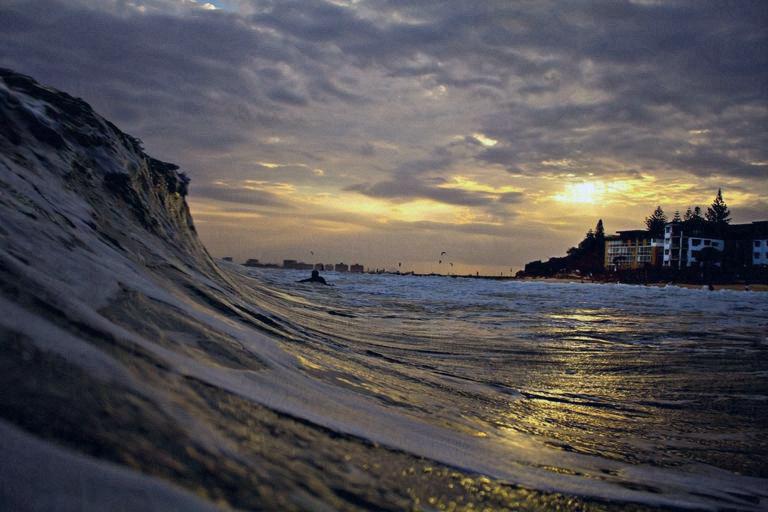}
        \caption[]%
        {{Coupled Optimization, $\epsilon=0.05$}}    
        \label{maps44}
    \end{subfigure}
    \caption[ Maps used to localize perturbations ]
    { The most promising configurations contrasted with the vanilla FGSM.} 
    \label{maps}
\end{figure}

\begin{table}[t]
\begin{center}
\begin{tabularx}{\linewidth}{llll}
\toprule
\textbf{Methods}	& $\epsilon$ &\textbf{Top-1 Acc.}$\downarrow$ & \textbf{NIMA Score}$\uparrow$\\

\midrule
\multirow{2}{*}{Salient Defence}& 0.01 & 	0.937 &	4.63\\
& 0.05 &	0.735 &	4.58 \\
\midrule
Salient Defence \& tilt-shift& 0.01 &	0.868&	\textbf{4.75} \\
\midrule
\multirow{2}{*}{Coupled Optimization}& 0.01 &	0.917&	4.63 \\
& 0.05 &	\textbf{0.458} &	4.54 \\
\midrule
Original Test Set & - & 1.0 & 4.64 \\
\bottomrule
\end{tabularx}
\end{center}
\caption{Results on MediaEval test set. Top-1 accuracy refers to the the prediction accuracy of the attack model (ResNet50 trained on  Places365-standard data set). The \textit{NIMA Score} column represents the average of the aesthetics scores.}
\label{tab:Results}
\end{table}

\section{Discussion and Outlook}
An obvious shortcoming of our salient defence algorithm is that saliency is subject to change after manipulations to the image. One way of improving this would be to predict the saliency of the perturbed image and reapply the modification on the original using this information. Also, the Sobel filter assigns gradients in the image such as that of the horizon as similar to edge-dense areas, resulting in a map where some flat areas are not obscured.
Furthermore, we have shown that NIMA is not reliable when assessing the corrupting quality of low-level noise such as FGSM perturbations. We believe that aesthetic algorithms trained for low-level cues would improve the efficacy of our coupled optimization approach.

\begin{acks}
This publication has emanated from research conducted with the financial support of Science Foundation Ireland (SFI) under grant number SFI/15/SIRG/3283 and SFI/12/RC/2289
\end{acks}

\bibliographystyle{ACM-Reference-Format}
\bibliography{sigproc} 


\begin{thebibliography}{00}


\ifx \showCODEN    \undefined \def \showCODEN     #1{\unskip}     \fi
\ifx \showDOI      \undefined \def \showDOI       #1{#1}\fi
\ifx \showISBNx    \undefined \def \showISBNx     #1{\unskip}     \fi
\ifx \showISBNxiii \undefined \def \showISBNxiii  #1{\unskip}     \fi
\ifx \showISSN     \undefined \def \showISSN      #1{\unskip}     \fi
\ifx \showLCCN     \undefined \def \showLCCN      #1{\unskip}     \fi
\ifx \shownote     \undefined \def \shownote      #1{#1}          \fi
\ifx \showarticletitle \undefined \def \showarticletitle #1{#1}   \fi
\ifx \showURL      \undefined \def \showURL       {\relax}        \fi
\providecommand\bibfield[2]{#2}
\providecommand\bibinfo[2]{#2}
\providecommand\natexlab[1]{#1}
\providecommand\showeprint[2][]{arXiv:#2}

\bibitem[\protect\citeauthoryear{Choi, Larson, Li, Li, Friedland, and
  Hanjalic}{Choi et~al\mbox{.}}{2017}]%
        {choi2017geo}
\bibfield{author}{\bibinfo{person}{Jaeyoung Choi}, \bibinfo{person}{Martha
  Larson}, \bibinfo{person}{Xinchao Li}, \bibinfo{person}{Kevin Li},
  \bibinfo{person}{Gerald Friedland}, {and} \bibinfo{person}{Alan Hanjalic}.}
  \bibinfo{year}{2017}\natexlab{}.
\newblock \showarticletitle{The geo-privacy bonus of popular photo
  enhancements}. In \bibinfo{booktitle}{{\em Proceedings of the 2017 ACM on
  International Conference on Multimedia Retrieval}}. ACM,
  \bibinfo{pages}{84--92}.
\newblock


\bibitem[\protect\citeauthoryear{Goodfellow, Shlens, and Szegedy}{Goodfellow
  et~al\mbox{.}}{2014}]%
        {goodfellow2014explaining}
\bibfield{author}{\bibinfo{person}{Ian~J Goodfellow}, \bibinfo{person}{Jonathon
  Shlens}, {and} \bibinfo{person}{Christian Szegedy}.}
  \bibinfo{year}{2014}\natexlab{}.
\newblock \showarticletitle{Explaining and harnessing adversarial examples}.
\newblock \bibinfo{journal}{{\em arXiv preprint arXiv:1412.6572\/}}
  (\bibinfo{year}{2014}).
\newblock


\bibitem[\protect\citeauthoryear{He, Zhang, Ren, and Sun}{He
  et~al\mbox{.}}{2016}]%
        {resnet}
\bibfield{author}{\bibinfo{person}{Kaiming He}, \bibinfo{person}{Xiangyu
  Zhang}, \bibinfo{person}{Shaoqing Ren}, {and} \bibinfo{person}{Jian Sun}.}
  \bibinfo{year}{2016}\natexlab{}.
\newblock \showarticletitle{Deep residual learning for image recognition}. In
  \bibinfo{booktitle}{{\em Proceedings of the IEEE conference on computer
  vision and pattern recognition}}. \bibinfo{pages}{770--778}.
\newblock


\bibitem[\protect\citeauthoryear{Jiang, Huang, Duan, and Zhao}{Jiang
  et~al\mbox{.}}{2015}]%
        {jiang2015salicon}
\bibfield{author}{\bibinfo{person}{Ming Jiang}, \bibinfo{person}{Shengsheng
  Huang}, \bibinfo{person}{Juanyong Duan}, {and} \bibinfo{person}{Qi Zhao}.}
  \bibinfo{year}{2015}\natexlab{}.
\newblock \showarticletitle{SALICON: Saliency in context}. In
  \bibinfo{booktitle}{{\em Proceedings of the IEEE conference on computer
  vision and pattern recognition}}. \bibinfo{pages}{1072--1080}.
\newblock


\bibitem[\protect\citeauthoryear{Linardos, Mohedano, Nieto, O'Connor, Giro-i
  Nieto, and McGuinness}{Linardos et~al\mbox{.}}{2019}]%
        {linardos2019simple}
\bibfield{author}{\bibinfo{person}{Panagiotis Linardos}, \bibinfo{person}{Eva
  Mohedano}, \bibinfo{person}{Juan~Jose Nieto}, \bibinfo{person}{Noel~E
  O'Connor}, \bibinfo{person}{Xavier Giro-i Nieto}, {and}
  \bibinfo{person}{Kevin McGuinness}.} \bibinfo{year}{2019}\natexlab{}.
\newblock \showarticletitle{Simple vs complex temporal recurrences for video
  saliency prediction}.
\newblock \bibinfo{journal}{{\em arXiv preprint arXiv:1907.01869\/}}
  (\bibinfo{year}{2019}).
\newblock


\bibitem[\protect\citeauthoryear{Liu and Zhao}{Liu and Zhao}{2018}]%
        {liu2018first}
\bibfield{author}{\bibinfo{person}{Zhuoran Liu} {and} \bibinfo{person}{Zhengyu
  Zhao}.} \bibinfo{year}{2018}\natexlab{}.
\newblock \showarticletitle{First Steps in Pixel Privacy: Exploring Deep
  Learning-based Image Enhancement against Large-Scale Image Inference.}. In
  \bibinfo{booktitle}{{\em MediaEval}}.
\newblock


\bibitem[\protect\citeauthoryear{Liu, Zhao, and Larson}{Liu
  et~al\mbox{.}}{2019}]%
        {mediaeval2019}
\bibfield{author}{\bibinfo{person}{Zhuoran Liu}, \bibinfo{person}{Zhengyu
  Zhao}, {and} \bibinfo{person}{Martha Larson}.}
  \bibinfo{year}{2019}\natexlab{}.
\newblock \showarticletitle{Pixel Privacy 2019: Protecting Sensitive Scene
  Information in Images}. In \bibinfo{booktitle}{{\em Working Notes Proceedings
  of the MediaEval 2019 Workshop}}.
\newblock


\bibitem[\protect\citeauthoryear{Murray, Marchesotti, and Perronnin}{Murray
  et~al\mbox{.}}{2012}]%
        {AVA}
\bibfield{author}{\bibinfo{person}{Naila Murray}, \bibinfo{person}{Luca
  Marchesotti}, {and} \bibinfo{person}{Florent Perronnin}.}
  \bibinfo{year}{2012}\natexlab{}.
\newblock \showarticletitle{AVA: A large-scale database for aesthetic visual
  analysis}. In \bibinfo{booktitle}{{\em 2012 IEEE Conference on Computer
  Vision and Pattern Recognition}}. IEEE, \bibinfo{pages}{2408--2415}.
\newblock


\bibitem[\protect\citeauthoryear{Sobel and Feldman}{Sobel and Feldman}{1968}]%
        {Sobel19683x3}
\bibfield{author}{\bibinfo{person}{Irwin Sobel} {and} \bibinfo{person}{Gary
  Feldman}.} \bibinfo{year}{1968}\natexlab{}.
\newblock \showarticletitle{A 3x3 isotropic gradient operator for image
  processing}.
\newblock \bibinfo{journal}{{\em a talk at the Stanford Artificial Project
  in\/}} (\bibinfo{year}{1968}), \bibinfo{pages}{271--272}.
\newblock


\bibitem[\protect\citeauthoryear{Talebi and Milanfar}{Talebi and
  Milanfar}{2018}]%
        {talebi2018nima}
\bibfield{author}{\bibinfo{person}{Hossein Talebi} {and}
  \bibinfo{person}{Peyman Milanfar}.} \bibinfo{year}{2018}\natexlab{}.
\newblock \showarticletitle{NIMA: Neural image assessment}.
\newblock \bibinfo{journal}{{\em IEEE Transactions on Image Processing\/}}
  \bibinfo{volume}{27}, \bibinfo{number}{8} (\bibinfo{year}{2018}),
  \bibinfo{pages}{3998--4011}.
\newblock


\bibitem[\protect\citeauthoryear{Zhou, Lapedriza, Khosla, Oliva, and
  Torralba}{Zhou et~al\mbox{.}}{2017}]%
        {zhou2017places}
\bibfield{author}{\bibinfo{person}{Bolei Zhou}, \bibinfo{person}{Agata
  Lapedriza}, \bibinfo{person}{Aditya Khosla}, \bibinfo{person}{Aude Oliva},
  {and} \bibinfo{person}{Antonio Torralba}.} \bibinfo{year}{2017}\natexlab{}.
\newblock \showarticletitle{Places: A 10 million image database for scene
  recognition}.
\newblock \bibinfo{journal}{{\em IEEE transactions on pattern analysis and
  machine intelligence\/}} \bibinfo{volume}{40}, \bibinfo{number}{6}
  (\bibinfo{year}{2017}), \bibinfo{pages}{1452--1464}.
\newblock


\end{thebibliography}

\end{document}